\documentclass[10pt,twocolumn,letterpaper]{article}
\usepackage{iccv}
\usepackage{times}
\usepackage{epsfig}
\usepackage{graphicx}
\usepackage{amsmath}
\usepackage{amssymb}
\usepackage{flexisym}
\usepackage{breqn}
\usepackage{subfigure}
\usepackage{soul}
\usepackage{enumitem}
\usepackage[linesnumbered,ruled]{algorithm2e}
\usepackage[table]{xcolor}
\usepackage{colortbl}
\definecolor{Gray}{gray}{0.85}
\usepackage{array,multirow,graphicx}
\usepackage{xfrac}
\usepackage{setspace}

\newcommand{\tr}{{\rm tr}}
\newcommand{\range}{{\sl range}}
\newcommand{\lspan}{{\sl span}}

\iccvfinalcopy 

\def\iccvPaperID{1339}

\begin{document}

\title{A Novel Space-Time Representation on the Positive Semidefinite Cone \\for Facial Expression Recognition}

\author{Anis~Kacem$^1$, Mohamed~Daoudi$^1$, Boulbaba~Ben~Amor$^1$, and Juan~Carlos~Alvarez-Paiva$^2$ \\
\\
$^1$IMT Lille Douai, Univ. Lille, CNRS, UMR 9189 -- CRIStAL -- \\Centre de
Recherche en Informatique Signal et Automatique de Lille, F-59000 Lille, France\\
$^2$Univ. Lille, CNRS, UMR 8524, Laboratoire Paul Painlev\'e, F-59000 Lille, France.
}

\maketitle
\thispagestyle{empty}

\begin{abstract}
   In this paper, we study the problem of facial expression recognition using a novel space-time geometric representation. We describe the temporal evolution of facial landmarks as parametrized trajectories on the Riemannian manifold of positive semidefinite matrices of fixed-rank. Our representation has the advantage to bring naturally a second desirable quantity when comparing shapes -- the spatial covariance -- in addition to the conventional affine-shape representation. We derive then geometric and computational tools for rate-invariant analysis and adaptive re-sampling of trajectories, grounding on the Riemannian geometry of the manifold. Specifically, our approach involves three steps: 1) facial landmarks are first mapped into the Riemannian manifold of positive semidefinite matrices of rank 2, to build time-parameterized trajectories; 2) a temporal alignment is performed on the trajectories, providing a geometry-aware (dis-)similarity measure between them; 3) finally, pairwise proximity function SVM (ppfSVM) is used to classify them, incorporating the latter (dis-)similarity measure into the kernel function. We show the effectiveness of the proposed approach on four publicly available benchmarks (CK+, MMI, Oulu-CASIA, and AFEW). The results of the proposed approach are comparable to or better than the state-of-the-art methods when involving only facial landmarks.
\end{abstract}

\section{Introduction}
In recent years, Automated Facial Expression Recognition (AFER) has aroused considerable interest \cite{Corneanu2016PAMI}. Earlier literature mostly focused on static faces grounding on either shape (geometry) or appearance features. Recently, there have been a general shift to exploit the dynamics (motion) in facial videos \cite{ElaiwatBB16,JungLYPK15,LiuSWC14}, as conveying an expression is obviously a temporal process. In particular, advances in landmarks detection \cite{AsthanaZCP14,ShenZCKTP15,XiongT13} have opened the door to accurate geometry-driven approaches. Besides, it has been stated that in unconstrained scenario, geometric features outperform appearance features \cite{Kossaifi2017}. However, analyzing temporal shape features brings new challenges -- (1) which suitable representation to facial shape analysis under rigid transformations due to changes in head position and orientation? (2) Which temporal representation for modeling the dynamic of facial expression? (3) How to compare and classify temporal sequences for the purpose of facial expression recognition? To tackle these challenges, we introduce in this work a comprehensive geometric framework which involves the temporal evolution of facial landmarks. Our framework incorporates a novel shape representation using Gramian matrices derived from centered facial landmark configurations and its extension to time-parametrized trajectories on the positive semidefinite cone. We use then appropriate tools to compare and classify trajectories in a rate-invariant fashion, grounding on the geometry of the manifold of interest. 


\section{Related Work}  
In the \textit{appearance-based (A)} category, first works extend conventional local features such us SIFT, LBP, and HOG to suit video-based data, giving rise to 3D SIFT \cite{ScovannerAS07}, LBP-TOP \cite{zhao2007dynamic}, and 3D HOG \cite{KlaserMS08}. In \cite{LiuSWC14}, the authors exploit the dynamics of facial expressions and propose a semantics-aware representation. They model a video clip as a Spatio-Temporal Manifold (STM) spanned by local spatio-temporal features called \textit{Expressionlets} built from low-level appearance features. These features are based on clustering cuboids of pre-defined sizes extracted from facial sequences in order to model the manifold of facial expression variations. A temporal alignment among STMs is performed to allow a rate-invariant analysis of facial expressions. Deep Networks based on appearance features have been recently applied on facial image sequences for the purpose of AFER.
Elaiwat \etal \cite{ElaiwatBB16} propose a restricted Boltzmann machine (RBM) network that, unlike typical deep models, is shallow and therefore easier to optimize. The key property of the RBM network is to disentangle expression-related image transformations from transformations that are not related to the expressions. Despite their investigation in expression recognition, deep networks are less effective if trained with small datasets \cite{Valstar2010idhas}.
To overcome this limitation, Jung \etal exploit two temporal features from the appearance and the geometry (landmarks) to train two deep networks termed respectively DTAN and DTGN \cite{JungLYPK15}. They are then combined using a joint fine-tuning method to give rise to the DTAGN. Finally, it has been shown in \cite{taigman2014deepface} that face analysis using deep networks is sensitive to pose variations and often requires a face alignment step. As far as the \textit{geometry-based (G)} approaches are concerned, in \cite{JainHA11}, the authors propose a probabilistic method to capture the subtle motions within expressions using Latent-Dynamic Conditional Random Fields (LDCRFs) on both geometric and appearance features. They illustrate experimentally that variations in shape are much more important than appearance for AFER. In another work, Wang \etal \cite{Wang2013CVPR} introduce a unified probabilistic framework based on an interval temporal Bayesian network (ITBN) built from the movements of specific geometric points detected on the face along a sequence.
Recently, shape trajectory-based methods showed their effectiveness in many temporal pattern recognition tasks, especially in action recognition \cite{anirudh2016elastic,bagheri2016support,Boulbaba2016PAMI, DevanneIEEESC2015,vemulapalli2016rolling}. Taheri \etal \cite{Taheri2001FG} propose an affine-invariant shape representation on the Grassmann manifold $\mathcal{G}(2,n)$ \cite{BegelforW06} and model the dynamic of facial expression by parametrized trajectories on this manifold. Geodesic velocities between facial shapes are then used to capture the facial deformations. The classification was achieved using LDA followed by SVM.

\begin{figure*}[!ht]
\centering
\includegraphics[width = .86\linewidth]{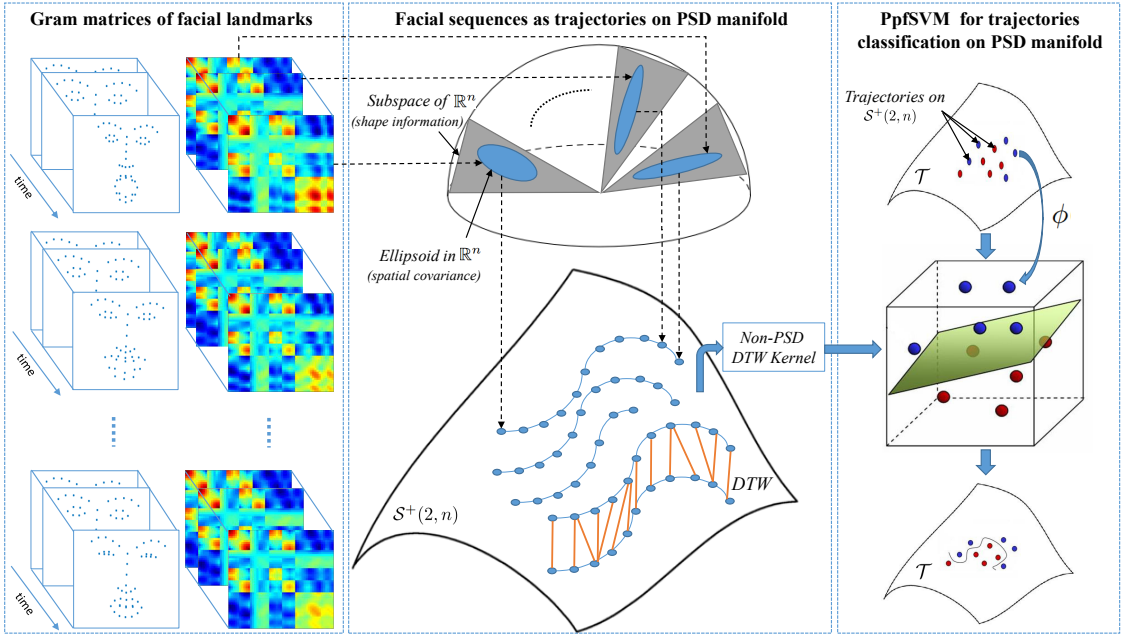}
\caption{Overview of the proposed approach -- After automatic landmark detection for each frame of the video, the Gram matrices are computed to build trajectories on $\mathcal{S}^+(2,n)$. A moving shape is hence assimilated to an ellipsoid traveling along $2$-dimensional subspaces of $\mathbb{R}^n$ with $d_{\mathcal{S}^+}$ to compare static ellipsoids. Dynamic Time Warping (DTW) is then used to align and compare trajectories in a rate-invariant manner. Finally, the ppfSVM is used on these trajectories as expression classifier.
\label{Fig:AppOverview}
}
\end{figure*}

From the discussion above, we propose a novel shape representation invariant to rigid motions by embedding shapes into a Positive Semidefinite Riemannian manifold. Facial expression sequences are then viewed as trajectories on this manifold. To compare and classify these trajectories, we propose a variant of SVM that takes into account the nonlinearity of this space. The full approach is illustrated in Fig.\ref{Fig:AppOverview}. In summary, the main contributions of this paper are:

\begin{itemize}[noitemsep,nolistsep, leftmargin=*]
\item  A novel static shape representation based on computing the Gramian matrix from centered landmark configurations, as well as a comprehensive study of the Riemannian geometry of the space of representations (called the cone of Positive Semidefinite $n\times 2$ matrices of fixed-rank $2$). Despite the large use of these matrices in several research fields, to the best of our knowledge, this is the first application in static and dynamic shape analysis.
\item   A temporal extension of the representation via parametrized trajectories in the underlying Riemannian manifold, with associated computational tools for temporal alignment and adaptive re-sampling of trajectories.
\item  The classification of trajectories based on pairwise proximity function SVM (ppfSVM) grounding on pairwise (dis-)similarity measures between them, with respect to the metric of the underlying manifold.
\item  Extensive experiments and baselines on four publicly datasets and a comparative study with existing literature, which demonstrates the competitiveness of the approach.  
\end{itemize}

The rest of the paper is organized as following. In section~\ref{sec:2}, we study the Riemannian geometry of the Positive Semidefinite manifold. In section~\ref{sec:3}, we adopt a temporal extension of the representation via time-parametrized trajectories in the manifold, with the definition of relevant geometric tools for temporal registration and trajectory re-sampling. Section~\ref{sec:4} states the classification approach based on a variant of the standard SVM associated to a closeness between the trajectories. Experimental results and discussions are reported in section~\ref{sec:5}. In section~\ref{sec:6}, we conclude and draw some perspectives of the work.

\section{Shape and Trajectory Representations}
\label{sec:2}

Let us consider $\{Z_0,\ldots, Z_N\}$ an arbitrary sequence of landmark configurations. Each configuration $Z_{i}$ $(0 \leq i\leq N)$ is an $n \times 2$ matrix $\left[ (x_1,y_1), (x_2,y_2),\ldots, (x_n,y_n) \right]^T$
of rank $2$ encoding the positions of $n$ distinct points on the plane: $p_1 = (x_1,y_1), \ldots , p_n = (x_n,y_n)$. We are interested in studying such sequences or curves of landmark configurations up to Euclidean motions of the plane. In what follows, we will first study static observations representation, then adopt a time-parametrized representation for a temporal analysis. 

As a first step, we seek a shape representation that is invariant up to Euclidean  transformations (rotation and translation). Arguably, the most natural choice is the matrix of pairwise distances between the landmark points of the same shape augmented by the distances from all landmarks to the center of mass $p_0 =(\bar{x},\bar{y})$. Since we are dealing with Euclidean distances, it will turn out to be more convenient to consider the matrix of the squares of these distances. Also note that by subtracting the center of mass from the coordinates of the landmarks, these can be considered as \textit{centered}: the center of mass is always at the origin. From now on we will assume $p_0 =(\bar{x},\bar{y}) = (0, 0)$. With  this provision, the augmented pairwise square-distance matrix $\mathcal{D}$ takes the form,
$$
\centering
\mathcal{D} :=
\begin{pmatrix}
0 & \|p_1\|^2 &\cdots & \|p_n\|^2 \\
\|p_1\|^2 & 0 & \cdots & \|p_1 - p_n\|^2 \\
\vdots & \vdots & \vdots & \vdots  \\
\|p_n\|^2 & \|p_n - p_1\|^2 & \cdots & 0 \\
\end{pmatrix},
$$
where $p_i = (x_i,y_i)$ for all $ 1 \leq i\leq n$. As usual, $\| \cdot \|$ denotes the norm associated to the  $l^2$-inner product $\langle \cdot , \cdot \rangle$.

A key observation is that the matrix $\mathcal{D}$ can be easily read from the 
$n \times n$ Gram matrix $G:= ZZ^T$. Indeed, the  entries of $G$ are the 
pairwise inner products of the points $p_1, \ldots, p_n$,
\begin{equation}
\label{eq:gram}
G=ZZ^T = \langle  p_i , p_j\rangle \ \  1 \leq i, j \leq n \ ,
\end{equation}
and the equality 
\begin{equation}
\mathcal{D}_{ij} = 
\langle  p_i , p_i \rangle 
- 2 \langle  p_i , p_j \rangle
+ \langle  p_j , p_j \rangle  \ \ (0 \leq i, j \leq n),
\end{equation}
establishes a linear equivalence between the set of $n \times n$ Gram matrices and augmented square-distance $(n+1) \times (n+ 1)$ matrices of distinct points on the plane. On the other hand, Gram matrices of the form
$ZZ^T$, where $Z$ is an $n \times 2$ matrix of rank $2$, are characterized as $n \times n$ positive semidefinite matrices of rank $2$ (for a detailed account of the relation between positive semidefinte matrices, Gram matrices, and square-distance matrices, we refer the reader to Section 6.2.1 of the book \cite{deza2009geometry}). Conveniently for us, the Riemannian geometry of the space of these matrices, called the positive semidefinite cone $\mathcal{S}^+(2,n)$, was studied in \cite{Bonnabel2009SIAM,faraki2016image,meyer2011regression,vandereycken2009embedded}.

An alternative shape representation considered in \cite{BegelforW06} and \cite{Taheri2001FG} associates to each configuration $Z$
the two-dimensional subspace $\lspan(Z)$ spanned by its columns. This representation, which exploits the well-known geometry of the Grassmann manifold $\mathcal{G}(2,n)$ of two-dimensional subspaces in $\mathbb{R}^n$, is invariant under {\it all} invertible linear transformations. 
By fully encoding the set of all mutual distances between landmark points, the Euclidean shape representation proposed in this paper supplements the affine shape representation with the knowledge of the $2 \times 2$ covariance matrix for the centered landmarks. This leads to considerable improvements in the results of the conducted facial expression recognition experiments.
\subsection{Riemannian geometry of $\mathcal{S}^+(2,n)$}
Given an $n \times 2$ matrix $Z$ of rank two, its polar decomposition $Z = UR$ 
with $R = (Z^T Z)^{1/2}$ allows us to write the Gram matrix $ZZ^T$ as $UR^2U^T$. Since the columns of the matrix $U$ are orthonormal, this decomposition defines a map
\begin{alignat*}{2}
\Pi : & {V}_{n,2} \times  \mathcal{P}_2 \rightarrow \mathcal{S}^+(2,n) \\
  &(U,R^2)\mapsto UR^2U^T 
\end{alignat*}
from the product of the Stiefel manifold ${V}_{n,2}$ and the cone of $2 \times 2$ positive definite matrices $\mathcal{P}_2$ to the manifold $\mathcal{S}^+(2,n)$ of $n \times n$ positive semidefinite matrices of rank two. The map $\Pi$ defines a principal fiber bundle over $\mathcal{S}^+(2,n)$ with fibers
$$
\Pi^{-1}(UR^2U^T) = \{(UO, O^TR^2O) : O \in \mathcal{O}(2) \} \ ,
$$ where $\mathcal{O}(2)$ is the group of $2 \times 2$ orthogonal matrices. 
Bonnabel and Sepulchre \cite{Bonnabel2009SIAM} use this map and the geometry of the {\it structure space} ${V}_{n,2} \times  \mathcal{P}_2$ to introduce a Riemannian metric on $ \mathcal{S}^+(2,n)$
and study its geometry.

\subsection{Tangent space and Riemannian metric}

The tangent space $T_{(U,R^2)}(V_{n,2} \times \mathcal{P}_2)$ consists of pairs $(M,N)$, where $M$ is a $n \times 2$ matrix satisfying $M^TU + U^TM = 0$ and $N$ is any 
$2 \times 2$ symmetric matrix. Bonnabel and Sepulchre define a {\it connection} (see \cite[p.~63]{Kobayashi-Nomizu:1963}) on the principal bundle $\Pi : {V}_{n,2} \times  \mathcal{P}_2 \rightarrow \mathcal{S}^+(2,n)$ by setting the
horizontal subspace $\mathcal{H}_{(U,R^2)}$  at the point $(U,R^2)$ to be the space of tangent vectors $(M,N)$ such that $M^TU = 0$ and $N$ is an arbitrary $2 \times 2$ symmetric matrix. They also define an inner product on $\mathcal{H}_{(U,R^2)}$: given two tangent vectors $A=(M_1, N_1)$ and $B= (M_2, N_2)$ on $\mathcal{H}_{(U,R^2)}$, set
\begin{equation}
\centering
\langle (A,B) \rangle_{\mathcal{H}_{U,R^2}}=\tr(M_1^TM_2) + k \ \tr(N_1R^{-2}N_2R^{-2}),
\label{eq:Rmetrics}
\end{equation}
where $k > 0$ is a real parameter. 

It is easily checked that the action of the group of $2 \times 2$ orthogonal matrices on the fiber $\Pi^{-1}(UR^2U^T)$ sends horizontals to horizontals isometrically. It follows that the inner product on 
$T_{UR^2U^T}\mathcal{S}^+(2,n)$ induced from that of $\mathcal{H}_{(U,R^2)}$ via the linear isomorphism $D\Pi$ is independent of the choice of point $(U,R^2)$ projecting onto $UR^2U^T$. This procedure defines a Riemannian metric on $\mathcal{S}^+(2,n)$ for which the natural projection 
\begin{alignat*}{2}
  \rho : \ &  \mathcal{S}^+(2,n) \rightarrow \mathcal{G}(2,n) \\
  &G\mapsto \range(G) 
\end{alignat*}
is a Riemannian submersion. This allows us to relate the geometry of $\mathcal{S}^+(2,n)$ with that of the Grassmannian $\mathcal{G}(2,n)$.

Recall that the geometry of the Grassmannian $\mathcal{G}(2,n)$ is easily described by using the
map
$$
\lspan : V_{n,2} \rightarrow \mathcal{G}(2,n)
$$
that sends an $n \times 2$ matrix with orthonormal columns $U$ to their span $\lspan(U)$. Given two subspaces $\mathcal{U}_1 = \lspan(U_1)$ and $\mathcal{U}_2=\lspan(U_2) \in \mathcal{G}(2,n)$, the geodesic curve connecting them is
\begin{dmath}
\label{eq:GeoGrass}
\lspan(U(t))= \lspan(U_{1}\cos(\Theta t)+M\sin(\Theta t)),
\end{dmath}
where $\Theta=diag(\theta_1,\theta_2)$ is a $2 \times 2$ diagonal matrix formed by the \textit{principal angles} between $\mathcal{U}_1$ and $\mathcal{U}_2$, while the matrix $M$ is given by the formula $M=(I_n-U_{1}U_1^T)U_2 F$, with $F$ being the pseudoinverse $diag(\sin(\theta_1),\sin(\theta_2))$. The Riemannian distance between 
$\mathcal{U}_1$ and $\mathcal{U}_2$ is given by 
\begin{equation}
d^2_{\mathcal{G}}(\mathcal{U}_1,\mathcal{U}_2)=\|\Theta\|^2_F.
\label{eq:distGrass}
\end{equation}

\subsection{Pseudo-geodesics and closeness in $\mathcal{S}^+(2,n)$}

Bonnabel and Sepulchre (\cite{Bonnabel2009SIAM}) define the \textit{pseudo-geodesic} connecting 
two matrices $G_1 = U_1R_1^2U_1^T$ and $G_2 =  U_2R_2^2U_2^T$ in $\mathcal{S}^+(2,n)$ as the curve
\begin{equation}
\mathcal{C}_{G_1\to G_2}(t)=U(t)R^2(t)U^T(t), \forall t \in [0,1],
\label{eq:geopsd}
\end{equation}
where $R^2(t)=R_{1}\exp(t\log R_{1}^{-1}R_{2}^{2}R_{1}^{-1})R_{1}$ is a geodesic in $\mathcal{P}_2$ and $U(t)$ is the geodesic in $\mathcal{G}(2,n)$ given by Eq.(\ref{eq:GeoGrass}). They also define the {\it closeness} between $G_1$ and $G_2$, $d_{\mathcal{S}^+}(G_1,G_2)$, as the square of the length of this curve:
\begin{dmath}
\label{eq:closeness}
d_{\mathcal{S}^+}(G_1,G_2)=\|\Theta \|_F^2+k\|\log R_{1}^{-1}R_{2}^{2}R_{1}^{-1} \|_F^2
=d_{\mathcal{G}}^2(\lspan(U_{1}),\lspan(U_{2}))+kd_{\mathcal{P}_2}^2(R_{1}^2,R_{2}^2) \ .
\end{dmath}

\begin{figure}[h]
  \centering
   \includegraphics[width=\linewidth]{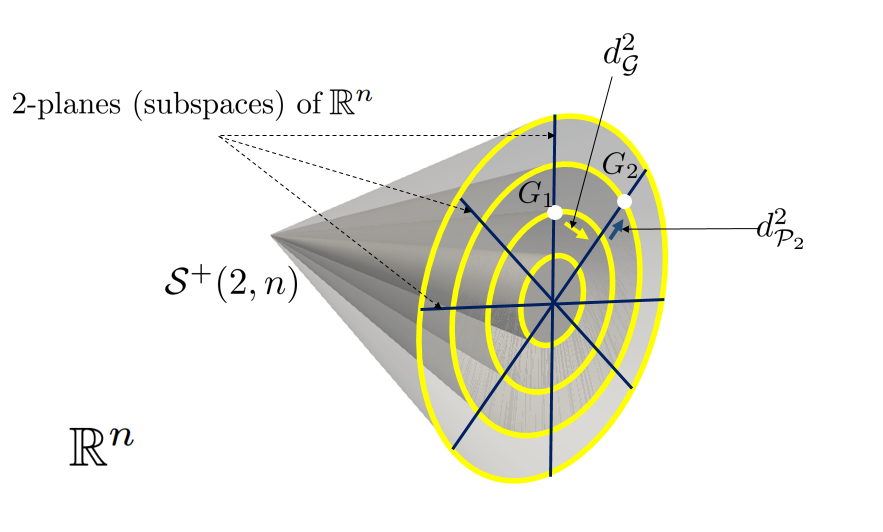}
  \caption{A pictorial representation of the positive semidefinite cone $\mathcal{S}^+(2,n)$. Viewing matrices $G_1$ and $G_2$ as ellipsoids in $\mathbb{R}^{n}$, the closeness consists of two contributions: $d^2_{\mathcal{G}}$ (squared Grassmann distance) and $d^2_{\mathcal{P}_2}$ (squared Riemannian distance in $\mathcal{P}_2$).}
   \label{Fig:Cone}
\end{figure}

The closeness $d_{\mathcal{S}^+}$ consists of two independent contributions, the square of the distance  $d_{\mathcal{G}}(\lspan(U_1),\lspan(U_2))$ between the two associated subspaces and the 
square of the distance $d_{\mathcal{P}_2}(R_1^2,R_2^2)$ on the positive cone $\mathcal{P}_2$ (Fig.\ref{Fig:Cone}). Note that $\mathcal{C}_{G_1\to G_2}$ is not necessarily a geodesic and therefore, the closeness $d_{\mathcal{S}^+}$ is not a true Riemannian distance. From the viewpoint of
the landmark configurations $Z_1$ and $Z_2$, with $G_1 = Z_1 Z_1^T$ and $G_2 = Z_2 Z_2^T$, the closeness encodes the distances measured between the affine shapes $\lspan(Z_1)$ and $\lspan(Z_2)$ in $\mathcal{G}(2,n)$ and between their spatial covariances in $\mathcal{P}_2$. Indeed, the  spatial covariance of $Z_i$ $(i = 1, 2)$ is the $2 \times 2$ symmetric positive definite matrix 
\begin{equation}
C = \frac{ Z_i^TZ_i}{n-1}=\frac{(U_iR_i)^T(U_iR_i)}{n-1}=\frac{R_i^2}{n-1}.
\label{eq:covariance}
\end{equation}
 
The weight parameter $k$ controls the relative contribution of these two informations. Note that for $k=0$ the distance on $\mathcal{S}^+(2,n)$ collapses to the distance on $\mathcal{G}(2,n)$. Nevertheless, the authors in \cite{Bonnabel2009SIAM} recommend choosing small values for this parameter. The conducted experiments for expression recognition reported in section~\ref{sec:5} are in accordance with this recommendation.
  
For more details about the geometry of the Grassmannians $\mathcal{G}(2,n)$ and the positive define cone ${\mathcal{P}_2}$, readers are referred  to \cite{absil2004riemannian,BegelforW06,Bonnabel2009SIAM,Pennec2006IJCV}. 

\section{Modeling Facial Expressions as Trajectories in $\mathcal{S}^+(2,n)$} 
\label{sec:3}

We are able to compare static shape representations based on their 
Gramian representation $G$, the induced space, and closeness introduced in the previous section. We need a natural and effective extension to study their temporal evolution. Following \cite{Boulbaba2016PAMI,Taheri2001FG,vemulapalli2014human}, we define curves $\beta_G~:I~\rightarrow~\mathcal{S}^+(2,n)$ (I denotes the time domain, \eg $[0,1]$) to model the spatio-temporal evolution of elements on $\mathcal{S}^+(2,n)$. Given a sequence of shapes $\{Z_0, \ldots, Z_N\}$ represented by their corresponding Gram matrices $\{G_0,\ldots,G_N\}$ in $\mathcal{S}^+(2,n)$, the corresponding curve is the trajectory of the point $\beta_G(t)$ on $\mathcal{S}^+(2,n)$, when $t$ ranges in $[0,1]$. These curves are obtained by connecting all successive Gramian representations of shapes $G_i$  and $G_{i+1}$, $0 \leq i\leq N-1$ , by pseudo-geodesics in $\mathcal{S}^+(2,n)$.

\subsection{Temporal alignment and analysis}

The execution rate of facial expressions is often arbitrary and that results in different parameterizations of corresponding trajectories. This parameterization variability distorts the comparison measures of these trajectories. Given $m$ trajectories $\{\beta_G^{1}, \beta_G^{2}, \ldots, \beta_G^{m} \}$ on $\mathcal{S}^+(2,n)$,  we are interested in finding functions $\gamma_i$ such that the $\beta_G^{i}(\gamma_i(t))$ are matched optimally for all $t \in [0,1]$. In other words, two curves $\beta_G^{1}(t)$ and $\beta_G^{2}(t)$ represent the same trajectories if their images are the same. This happens if, and only if, $\beta_G^{2} = \beta_G^{1} \circ \gamma$, where $\gamma$ is a re-parameterization of the interval $[0,1]$. 
The problem of temporal alignment is turned to find an optimal warping function $\gamma^{\star}$ according to,
 \begin{equation}
 \gamma^{\star}=\arg\min_{\gamma \in \Gamma} \int_0^1 d_{\mathcal{S}^+}(\beta_G^{1}(t),\beta_G^{2}(\gamma(t))) \, \mathrm{d}t \ ,
 \label {eq:DTW}
 \end{equation}
where $\Gamma $ denotes the set of all monotonically-increasing functions $\gamma : [0,1]\to [0,1]$. The most commonly used method to solve such optimization problem is the Dynamic Time Warping (DTW) algorithm. Note that accommodation of the DTW algorithm to the manifold-values sequences can be achieved with respect to an appropriate metric defined on the underlying manifold $\mathcal{S}^+(2,n)$. Having the optimal re-parametrization function $\gamma^{\star}$, one can define a (dis-)similarity measure between two trajectories allowing a rate-invariant comparison: 
 \begin{equation}
 d_{DTW}(\beta_G^{1},\beta_G^{2})=\int_0^1 d_{\mathcal{S}^+}(\beta_G^{1}(t),\beta_G^{2}(\gamma^{\star}(t))) \, \mathrm{d}t . 
  \label {eq:comptraj}
 \end{equation}
 
From now, we shall use $d_{DTW}(.,.)$ to compare trajectories in our manifold of interest $\mathcal{S}^+(2,n)$.

 \subsection{Adaptive re-sampling of trajectories}

One difficulty in video analysis is to catch the most relevant frames and focus on them. In fact, it is relevant to reduce the number of frames when no motion happened and in the same time "introduce" new frames, otherwise. Our geometric framework provides tools to do so. In fact, interpolation between successive frames could be achieved using pseudo-geodesics defined in Eq.(\ref{eq:geopsd}), while their length (closeness defined in Eq.(\ref{eq:closeness})) expresses the magnitude of the motion. Accordingly, we have designed an adaptive re-sampling tool that is able to increase/decrease the number of samples in a fixed time interval according to their relevance, with respect to the geometry of the underlying manifold $\mathcal{S}^+(2,n)$. Relevant samples are identified by a relatively low closeness $d_{\mathcal{S}^+}$ to the previous frame, while irrelevant ones correspond to a higher closeness level. Here, the down-sampling is performed by removing irrelevant shapes. In turn, the up-sampling is possible by interpolating between successive shape representations in $\mathcal{S}^+(2,n)$ using pseudo-geodesics. 

More formally, given a trajectory $\beta_G(t)_{t=0,1,\ldots,N}$ on $\mathcal{S}^+(2,n)$, for each sample $\beta_G(t)$ we compute the closeness to the previous sample, \ie $d_{\mathcal{S}^+}(\beta_G(t),\beta_G(t-1))$. If the value is below a defined threshold $\zeta_1$, current sample is simply removed from the trajectory. In contrast, if the distance exceeds a second threshold $\zeta_2$, new samples (shapes) generated from the pseudo-geodesic curve connecting $\beta_G(t)$ to $\beta_G(t-1)$ are inserted in the trajectory.

\section{Classification of Trajectories in $\mathcal{S}^+(2,n)$}
\label{sec:4}

Our trajectory representation reduces the problem of facial sequences classification to trajectories classification in $\mathcal{S}^+(2,n)$. Let us consider $\mathcal{T}= \{ \beta_G~:[0,1]\rightarrow \mathcal{S}^+(2,n)\}$, the set of time-parametrized trajectories of the underlying manifold. Let $\mathcal{L} = \{(\beta_G^{1},y^{1}),\ldots,(\beta_G^{m},y^{m}) \}$ be the training set with class labels, where $\beta_G^{i} \in \mathcal{T}$ and $y^{i} \in {\cal Y}$, \eg ${\cal Y}=\{\textrm{(Ha)}, \textrm{(An)}\} $, such that $y^i = f(\beta_G^{i})$. The goal here is to find an approximation $h$ to $f$ such that $h:  \mathcal{T} \rightarrow \mathcal{L}$. In Euclidean spaces, any standard classifier (\eg standard SVM) may be a natural and appropriate choice to classify the trajectories. Unfortunately, this is no more suitable as the space $\mathcal{T}$ built from $\mathcal{S}^+(2,n)$ is non-linear. A function that divides the manifold is rather a complicated notion compared with the Euclidean space. In current literature, few approaches have been proposed to handle the nonlinearity of Riemannian manifolds \cite{JayasumanaHSLH15,Taheri2001FG,vemulapalli2014human,vemulapalli2016rolling}. These methods map the points on the manifold to a tangent space or to Hilbert space where traditional learning techniques can be used for classification. Mapping data to a tangent space only yields a first-order approximation of the data that can be distorted, especially in regions far from the origin of the tangent space. Moreover, iteratively mapping back and forth, \ie Riemannian Logarithmic and exponential maps, to the tangent spaces significantly increases the computational cost of the algorithm. Recently, some authors propose to embed a manifold in a high dimensional Reproducing Kernel Hilbert Space  (RKHS), where Euclidean geometry can be applied \cite{JayasumanaHSLH15}. The Riemannian kernels enable the classifiers to operate in an extrinsic feature space without computing tangent space and $\log$ and $\exp$ maps. Many Euclidean machine learning algorithms can be directly generalized to an RKHS, which is a vector space that possesses an important structure: the inner product. Such an embedding, however, requires a positive semi-definite kernel function, according to Mercer's theorem ~\cite{Scholkopf:2001:LKS:559923}.

Inspired by a recent work of \cite{bagheri2016support} for action recognition, we adopt the \textit{pairwise proximity function SVM} (ppfSVM) \cite{graepel1999classification,gudmundsson2008support}. PpfSVM requires the definition of a (dis-)simlarity measure to compare samples. In our case, it is natural to consider the $d_{DTW}$ defined in Eq.(\ref{eq:comptraj}) for such a comparison. This strategy involves the construction of inputs such that each trajectory is represented by its (dis-)similarity to all the trajectories in the dataset, with respect to $d_{DTW}$, and then apply a conventional SVM to this transformed data \cite{gudmundsson2008support}. The ppfSVM is related to the arbitrary kernel-SVM without restrictions on the kernel function \cite{graepel1999classification}.  

Given $m$ trajectories $\{\beta_G^{1}, \beta_G^{2}, \ldots, \beta_G^{m} \}$ in $\mathcal{T}$, we follow \cite{bagheri2016support} and define a proximity function $\mathcal{P}_{\mathcal{T}}: \mathcal{T} \times \mathcal{T} \rightarrow \mathbb{R}_+$ between two trajectories $\beta_G^{1}, \beta_G^{2} \in \mathcal{T} $ as following,  

\begin{equation}
\label{eq:proxf}
\mathcal{P}_{\mathcal{T}}(\beta_G^{1}, \beta_G^{2})=d_{DTW}(\beta_G^{1}, \beta_G^{2}).
\end{equation}

According to \cite{graepel1999classification}, there are no restrictions on the function $\mathcal{P}_{\mathcal{T}}$. For an input trajectory $\beta_G \in \mathcal{T}$, the mapping $\phi(\beta_G)$ is given by,  
\begin{equation}
\label{eq:mapping}
\phi(\beta_G)=[\mathcal{P}_{\mathcal{T}}(\beta_G,\beta_G^{1}),\ldots,\mathcal{P}_{\mathcal{T}}(\beta_G,\beta_G^{m}) ]^T.
\end{equation}

The obtained vector $\phi(\beta_G) \in \mathbb{R}^m$ is used to represent a sample trajectory $\beta_G \in \mathcal{T}$. Hence, the set of trajectories can be represented by a $m \times m$ matrix $P$, where $P(i,j)=\mathcal{P}_{\mathcal{T}}(\beta_G^{i},\beta_G^{j})$, $i,j \in \{1,\ldots,m\}$. Finally, a linear SVM is applied to this data representation. urther details on ppfSVM can be found in \cite{bagheri2016support,graepel1999classification,gudmundsson2008support}.

\section{Experimental Results}
\label{sec:5}

\begin{figure*}[h]
  \centering
   \includegraphics[width=\linewidth]{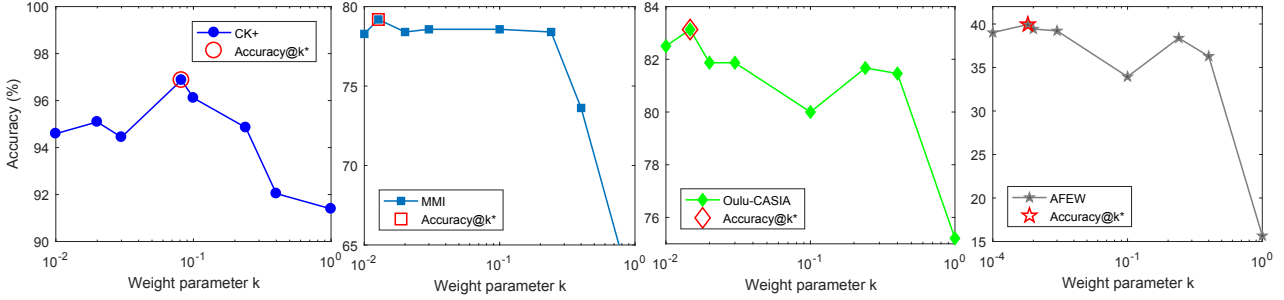}
  \caption{Accuracy of the proposed approach when varying the weight parameter $k$, on (left to right) CK+, MMI, Oulu-CASIA, and AFEW.}
   \label{Fig:Acc_vs_k}
\end{figure*}

To validate the proposed approach, we have conducted extensive experiments on four publicly available datasets -- CK+, MMI, Oulu-CASIA, and AFEW. We have followed experimental settings commonly used in recent works. Note that all our experiments are made once the facial landmarks are extracted using the method proposed in \cite{AsthanaZCP14} on CK+, MMI, and Oulu-CASIA datasets. On the challenging AFEW, we have considered the corrections provided in \footnote{ http://sites.google.com/site/chehrahome} after applying the same detector. 

\textbf{Cohn-Kanade Extended (CK+) database} \cite{LuceyCKSAM10} -- is one of the most popular datasets. It contains $123$ subjects and $593$ frontal image sequences of posed expressions. Among them, $118$ subjects are annotated with the seven labels -- anger (An), contempt (Co), disgust (Di), fear (Fe), happy (Ha), sad (Sa), and surprise (Su). Note that only the two first temporal phases of the expression, \ie neutral and onset (with apex frames), are present. Following the same settings of \cite{ElaiwatBB16,JungLYPK15}, we have performed $10$-fold cross validation experiment. The results are summarized in Table~\ref{tab:CM_CK+}. 

\begin{table}[h]
\centering
\caption{Confusion matrix of the proposed trajectory representation and classification on $\mathcal{S}^+(2,n)$ -- CK+ database.}
\label{tab:CM_CK+}
\footnotesize
\begin{tabular}{{c}|{c}{c}{c}{c}{c}{c}{c}}
   & An & Co   & Di   & Fe  & Ha   & Sa   & Su\\
 \hline 
An &\cellcolor{gray}\textbf{100} &5.55  &3.38  &0    &0     &3.57  &0\\

Co &0   &\cellcolor{gray}\textbf{83.35} &0     &0    &1.44  &0     &1.2\\

Di &0   &0     &\cellcolor{gray}\textbf{96.62} &0    &0     &0     &0\\

Fe &0   &0     &0     &\cellcolor{gray}\textbf{92}   &0     &0     &0\\

Ha &0   &5.55  &0     &8    &\cellcolor{gray}\textbf{98.56} &0     &0\\

Sa &0   &5.55  &0     &0    &0     &\cellcolor{gray}\textbf{96.43} &0 \\

Su &0   &0     &0     &0    &0     &0  &\cellcolor{gray}\textbf{98.8} 
\end{tabular}
\end{table}

Overall, the average accuracy is 96.87\%. When individual accuracy of (An), (Di), (Ha), and (Su) are high, recognizing (Co) and (Fe) is still challenging. Note that the accuracy of the trajectory representation on $\mathcal{G}(2,n)$, following the same pipeline is 2\% lower, which confirms the contribution of the covariance embedded to our shape representation.  

\textbf{MMI database} \cite{Valstar2010idhas} -- consists of $205$ image sequences with frontal faces of only $30$ subjects labeled with the six basic emotion labels. This database is different from the other databases; each sequence begins with a neutral facial expression and has a posed facial expression in the middle of the sequence. This ends with the neutral facial expression. The location of the peak frame is not provided as a prior information. Here again, the protocol used in \cite{ElaiwatBB16,JungLYPK15} was followed according to a $10$-fold cross-validation schema. The confusion matrix is reported in Table~\ref{tab:CM_MMI}. An average classification accuracy of 79.19\% is reported. Note that based on geometric features only, our approach grounding on both representations on $\mathcal{S}^+(2,n)$ and $\mathcal{G}(2,n)$ achieves competitive results with respect to the literature (see Table~\ref{tab:ARR_CK+_MMI}).

\begin{table}[h]
\centering
\caption{Confusion matrix of the proposed trajectory representation and classification on $\mathcal{S}^+(2,n)$ -- MMI database.}
\label{tab:CM_MMI}
\footnotesize
\begin{tabular}{{c}|{c}{c}{c}{c}{c}{c}}
 & An      &Di     & Fe      & Ha   & Sa   & Su\\
 \hline 
An &\cellcolor{gray}\textbf{76.66}     &9.37   &0     &0     &9.37  &0\\

Di &13.33   &\cellcolor{gray}\textbf{75}     &13.79    &2.44  &3.12     &0\\

Fe &0   &3.12   &\cellcolor{gray}\textbf{55.17}    &0  &9.37  &12.82 \\

Ha &0      &12.5  &0        &\cellcolor{gray}\textbf{97.56} &0  &0 \\

Sa &10     &0      &3.44     &0     &\cellcolor{gray}\textbf{71.87} &2.56 \\

Su &0      &0      &27.58    &0     &6.25  &\cellcolor{gray}\textbf{84.61} 
\end{tabular}
\end{table} 

\textbf{Oulu-CASIA database} \cite{ZhaoHTLP11} -- includes $480$ image sequences of $80$ subjects taken under normal illumination conditions. They are labeled with one of the six basic emotion labels. Each sequence begins with a neutral facial expression and ends with the apex of the expression. We adopt a $10$-fold cross validation schema similarly to \cite{JungLYPK15,LiuSWC14}. This time, the average accuracy is 83.13\%, hence 3\% higher than the Grassmann trajectory representation. This is the highest accuracy reported in literature (refer to Table~\ref{tab:ARR_AFEW}). 

\begin{table}[h]
\centering
\caption{Confusion matrix of the proposed trajectory representation and classification on $\mathcal{S}^+(2,n)$ -- Oulu-CASIA database.}
\label{CM_CASIA}
\footnotesize
\begin{tabular}{{c}|{c}{c}{c}{c}{c}{c}}
 & An & Di & Fe & Ha & Sa & Su\\
 \hline 
An &\cellcolor{gray}\textbf{81.25}  &15  &1.25  &0  &13.75 &0\\

Di &10  &\cellcolor{gray}\textbf{78.75}  &2.5  &0  &6.25 &0\\

Fe &1.25  &1.25  &\cellcolor{gray}\textbf{78.75}  &6.25 &3.75 &5 \\

Ha &1.25  &1.25  &3.75  &\cellcolor{gray}\textbf{91.25} &1.25 &1.25 \\

Sa &6.25  &3.75  &5  &2.5 &\cellcolor{gray}\textbf{75} &0 \\

Su &0  &0  &8.75  &0 &0 &\cellcolor{gray}\textbf{93.75}
\end{tabular}
\end{table}

\textbf{AFEW database} \cite{DhallGLG12} -- collected from movies showing close-to-real-world conditions, which depicts or simulates the spontaneous expressions in uncontrolled environment. According to the protocol defined in EmotiW'2013 \cite{DhallGJWG13}, the database is divided into three sets: training, validation, and test. The task is to classify each video clip into one of the seven expression categories (the six basic emotions plus the neutral). As the ground truth of test set is still unreleased, here we only report our results on the validation set for comparison with \cite{DhallGJWG13,ElaiwatBB16,LiuSWC14}. The average accuracy is 39.94\%. Unsurprisingly, the (Ne), (An), and (Ha) are better recognized over the rest. Despite their competitiveness with respect to recent literature, these results state clearly that the AFER "in-the-wild" is still a distant goal.   

\begin{table}[h]
\centering
\caption{Confusion matrix of the proposed trajectory representation and classification on $\mathcal{S}^+(2,n)$ -- AFEW database following the EmotiW'13 protocol \cite{DhallGJWG13}.}
\label{CM_CK+}
\footnotesize
\begin{tabular}{{c}|{c}{c}{c}{c}{c}{c}{c}}
 & An & Di & Fe & Ha & Ne & Sa & Su\\
 \hline 
An &\cellcolor{gray}\textbf{56.25}  &12.5  &30.43  &4.76 &7.93  &13.11 &26.08\\
Di &0  &\cellcolor{gray}\textbf{10}  &8.69  &4.76 &0  &6.55 &2.17 \\
Fe &7.81  &7.5  &\cellcolor{gray}\textbf{26.08}  &4.76 &7.93  &14.75 &19.56\\
Ha &10.93  &22.5  &10.87  &\cellcolor{gray}\textbf{66.66} &6.35 &11.47 &2.17 \\
Ne &9.37  &37.5  &10.87  &12.69 &\cellcolor{gray}\textbf{63.49}  &32.78 &30.43\\
Sa &10.93  &2.5  &6.52 &6.35  &11.11 &\cellcolor{gray}\textbf{18.03} &2.17  \\
Su &4.68  &7.5  &6.52  &0 &3.17 &3.27 &\cellcolor{gray}\textbf{17.39} 
\end{tabular}
\end{table}

In Fig.\ref{Fig:Acc_vs_k} we study the method's behavior when varying the parameter $k$ (of the closeness) defined in Eq.(\ref{eq:closeness}). Recall that $k$ serves to balance the contribution of the distance between covariance matrices living in $\mathcal{P}_2$ with respect to the Grassmann contribution $\mathcal{G}(2,n)$. The graphs report the method accuracy respectively on CK+, MMI, Oulu-CASIA, and AFEW. The optimal performances are achieved for the following values -- $k^*_{CK+}=0.081$, $k^*_{MMI}=0.012$, $k^*_{Oulu-CASIA}=0.014$, and $k^*_{AFEW}=0.001$. 

\begin{table}[h]
\centering
\caption{Overall accuracy (\%) on CK+ and MMI datasets. Here, $^{(A)}$: appearance; $^{(G)}$: geometry (or shape); $^{s}$: static; $^*$: Deep Learning based approach; last raw: ours}\label{tab:ARR_CK+_MMI}
  \footnotesize
\begin{tabular}{{l}|{c}|{c}}
\textbf{Method} & \textbf{CK+} & \textbf{MMI} \\
    \hline
    $^{(A)}$ 3D HOG \cite{KlaserMS08} (from \cite{JungLYPK15}) & 91.44 & 60.89\\ 
    $^{(A)}$ 3D SIFT \cite{ScovannerAS07} (from \cite{JungLYPK15}) & - & 64.39\\
    $^{(A)}$ Cov3D \cite{SaninSHL13} (from \cite{JungLYPK15}) & 92.3 & - \\
    \hline
    $^{(A)}$ MSR \cite{PtuchaTS11} (LOSO) & 91.4 & - \\ 
    $^{(A)}$ STM-ExpLet \cite{LiuSWC14} (10-fold)& \textbf{94.19} & \textbf{75.12}\\ 
    $^{(A)}$ CSPL \cite{ZhongLYLHM12} (10-fold) & 89.89 & 73.53 \\
    $^{(A)}$ F-Bases \cite{Sariyanidi2017} (LOSO) & \textbf{96.02} & \textbf{75.12}\\
    $^{(A)}$ ST-RBM \cite{ElaiwatBB16} (10-fold) & \textbf{95.66} & \textbf{81.63} \\
    $^{(A)}$ FaceNet2ExpNet$^{*,s}$ \cite{DingZC16} & \textbf{96.8} & - \\
    $^{(A)}$ 3DCNN-DAP \cite{LiuLSWC14} $^*$ (15-fold) & 87.9 & 62.2\\ 
    $^{(A)}$ DTAN \cite{JungLYPK15} $^*$ (10-fold)& 91.44 & 62.45 \\
    $^{(A+G)}$ DTAGN \cite{JungLYPK15} $^*$ (10-fold)& \textbf{97.25} & \textbf{70.24} \\
    \hline
    $^{(G)}$ DTGN \cite{JungLYPK15} $^*$ (10-fold) & 92.35 & 59.02\\
    $^{(G)}$ TMS \cite{JainHA11} (4-fold) & 85.84 & - \\ 
    $^{(G)}$ HMM \cite{Wang2013CVPR} (15-fold) & 83.5 & 51.5 \\
    $^{(G)}$ ITBN \cite{Wang2013CVPR} (15-fold) & 86.3 & 59.7 \\
    $^{(G)}$ Velocity on $\mathcal{G}(n,2)$\cite{Taheri2001FG} & 82.8 & -\\ 
    $^{(G)}$ traj. on $\mathcal{G}(2,n)$ (10-fold) & 94.25 $\pm$ 3.71 & 78.18 $\pm$ 4.87\\
    $^{(G)}$ \textbf{traj. on $\mathcal{S}^{+}(2,n)$ (10-fold)} & \textbf{96.87 $\pm$ 2.46} & \textbf{79.19 $\pm$ 4.62 }\\
  \end{tabular}
\end{table}


\textbf{Comparative study with the state-of-the-art.}
\label{sec:comp} In tables~\ref{tab:ARR_CK+_MMI} and \ref{tab:ARR_AFEW}, we compare our approach over the recent literature. Overall, our approach achieves competitive performances with respect to the most recent approaches. On CK+, we obtained the second highest accuracy. The ranked-first approach is DTAGN \cite{JungLYPK15}, in which two deep networks are trained on shape and appearance channels, then fused. Note that the geometry deep network (DTGN) achieved 92.35\%, which is much lower than ours. Furthermore, our approach outperforms the ST-RBM \cite{ElaiwatBB16} and the STM-ExpLet \cite{LiuSWC14}. On MMI dataset, our approach outperforms the DTAGN \cite{JungLYPK15} and the STM-ExpLet \cite{LiuSWC14}. However, it is behind ST-RBM \cite{ElaiwatBB16}. Note that the FaceNet2ExpNet \cite{DingZC16} is a pure static approach and is reported here as the state-of-the-art of static AFER. 

\begin{table}[!htb]
\centering
\caption{Overall accuracy on Oulu-CASIA and AFEW dataset (following the EmotiW'13 protocol~\cite{DhallGJWG13})}\label{tab:ARR_AFEW}
  \footnotesize
\begin{tabular}{{l}|{c}|{c}}
\textbf{Method} & \textbf{Oulu-CASIA} & \textbf{AFEW}  \\
     \hline
$^{(A)}$ HOG 3D \cite{KlaserMS08} & 70.63 & 26.90\\
$^{(A)}$ HOE \cite{WangQT13} & - & 19.54\\
$^{(A)}$ 3D SIFT \cite{ScovannerAS07} & 55.83 & 24.87\\
    \hline
$^{(A)}$ LBP-TOP \cite{ZhaoP07} & 68.13 & 25.13\\
$^{(A)}$ EmotiW \cite{DhallGJWG13} & - & 27.27\\
$^{(A)}$ STM \cite{LiuSWC14} & - & 29.19\\
$^{(A)}$ STM-ExpLet \cite{LiuSWC14} & 74.59 & 31.73\\
$^{(A+G)}$ DTAGN \cite{JungLYPK15} $^*$ (10-fold)&   \textbf{81.46} & - \\
$^{(A)}$ ST-RBM \cite{ElaiwatBB16} & - & \textbf{46.36}\\
\textbf{$^{(G)}$ traj. on $\mathcal{G}(2,n)$} & 80.0 $\pm$ 5.22 &  39.1 \\
\textbf{$^{(G)}$ traj. on $\mathcal{S}^{+}(2,n)$} & \textbf{83.13 $\pm$ 3.86} & \textbf{39.94} \\
  \end{tabular}
\end{table}

On Oulu-CASIA dataset, our approach shows a clear superiority to existing methods, in particular STM-ExpLet \cite{LiuSWC14} and DTGN \cite{JungLYPK15}. Elaiwat \etal \cite{ElaiwatBB16} do not report any results on this dataset. However, their approach achieved the highest accuracy on AFEW. Our approach is ranked second showing a superiority to remaining approaches on AFEW.  


\textbf{Baseline experiments.} Based on the results reported in table \ref{tab:baseline}, we discuss in this paragraph algorithms and their computational complexity with respect to baselines. 

\begin{table}[h]
  \centering
  \caption{Baseline experiments on CK+, MMI, and AFEW datasets.}\label{tab:baseline}
  \footnotesize

  \begin{tabular}{{l}|{c}|{c}}
\textbf{Distance} & \textbf{CK+} (\%) & \textbf{Time} (s) \\
        \hline
    Flat distance $d_{\mathcal{F}^+}$ & 93.78 $\pm$ 2.92 & 0.020 \\
    Distance $d_{\mathcal{P}_n}$ in $\mathcal{P}_n$  & 92.92 $\pm$ 2.45 & 0.816 \\
    Closeness $d_{\mathcal{S}^+}$ & \textbf{96.87 $\pm$ 2.46}  & 0.055 \\
    \hline
  \end{tabular}
  
\vspace{0.2cm}  
  \begin{tabular}{{l}|{c}|{c}|{c}}
\textbf{Temporal alignment} & \textbf{CK+ (\%)} & \textbf{MMI (\%)} & \textbf{Time} (s) \\
        \hline
    without DTW & 90.94 $\pm$ 4.23 & 66.93 $\pm$ 5.79 & 0.018 \\
    with DTW &  \textbf{96.87 $\pm$ 2.46} & \textbf{79.19 $\pm$ 4.62} & 0.055\\
    \hline
  \end{tabular}
  
  \vspace{0.2cm}  
  \begin{tabular}{{l}|{c}|{c}}
\textbf{Adaptive re-sampling} & \textbf{MMI (\%)} & \textbf{AFEW (\%)} \\
        \hline
    without resampling & 74.72 $\pm$ 5.34 & 36.81  \\
    with resampling &  \textbf{79.19 $\pm$ 4.62} & \textbf{39.94}\\
    \hline
  \end{tabular}
      \vspace{0.2cm}      
  
\begin{tabular}{{c}|{c}|{c}}
\textbf{Classifier} & \textbf{CK+ (\%)} &  \textbf{AFEW (\%)} \\
      \hline
  K-NN & 88.97 $\pm$ 6.14 & 29.77 \\
  ppf-SVM & \textbf{96.87 $\pm$ 2.46} &  \textbf{39.94} \\
  \hline
 \end{tabular}
\end{table}

We highlight firstly the superiority of the trajectory representation on $\mathcal{S}^+(2,n)$ over the Grassmannian (refer to Tables~\ref{tab:ARR_CK+_MMI} and \ref{tab:ARR_AFEW}). This is due to the contribution of the covariance part further to the conventional affine-shape analysis over the Grassmannian. Secondly, we have used different distances defined on $\mathcal{S}^+(2,n)$. Specifically, given two matrices $G_1$ and $G_2$ in $\mathcal{S}^+(2,n)$: (1) as proposed in \cite{wang2012covariance}, we used $d_{\mathcal{P}_n}$ that was defined in Eq.(\ref{eq:closeness}) to compare them through regularizing their ranks, \ie making them $n$ full-rank and considering them in $\mathcal{P}_n$ (the space of $n$-by-$n$ positive definite matrices), $d_{\mathcal{P}_n}(G_1,G_2)=d_{\mathcal{P}_n}(G_1+\epsilon I_n,G_2+\epsilon I_n)$; (2) we used the Euclidean flat distance $d_{\mathcal{F}^+}(G_1,G_2)=\|G_1-G_2\|_F$, where $\|.\|_F$ denotes the Frobenius-norm. The closeness $d_{\mathcal{S}^+}$ between two elements of $\mathcal{S}^+(2,n)$ defined in Eq.(\ref{eq:closeness}) is more suitable compared to the distance $d_{\mathcal{P}_n}$ and the flat distance $d_{\mathcal{F}^+}$. This demonstrates the importance of being faithful to the geometry of the manifold of interest. Another advantage of using $d_{\mathcal{S}^+}$ over $d_{\mathcal{P}_n}$ is the computational time as it involves $n$-by-$2$ and $2$-by-$2$ matrices instead of $n$-by-$n$ matrices. 

Table \ref{tab:baseline} reports the average accuracy when DTW is used or not in our pipeline on both CK+ and MMI datasets. It is clear from these experiments that a temporal alignment of the trajectories is a crucial step as an improvement of around $12\%$ is obtained on MMI and $6\%$ on CK+. The adaptive re-sampling tool is also analyzed. When it is involved in the pipeline, an improvement of around $5\%$ is achieved on MMI and $3\%$ on AFEW. 

In the last table, we compare the results of ppfSVM to a K-NN classifier for both CK+ and AFEW databases. 
Each test sequence is classified by a majority vote of its K-nearest neighbors using the (dis-)similarity measure defined in Eq.~\ref{eq:comptraj}. The number of nearest neighbors K to consider for each database is chosen by cross-validation. On CK+, we obtained an average accuracy of $88.97\%$ for $K=11$. On AFEW, we obtained an average accuracy of $29.77\%$ for $K=7$. These results are outperformed by ppfSVM classifier.

\section{Conclusion and Future Work} 
\label{sec:6}
We have proposed in this paper a geometric approach for effectively modeling and classifying dynamic facial sequences. Based on Gramian matrices derived from the facial landmarks, our representation consists of an affine-invariant shape representation and a spatial covariance of the landmarks. We have exploited the geometry of the space to define a closeness between static and dynamic (trajectory) representations. We have derived then computational tools to align, re-sample and compare these trajectories giving rise to a rate-invariant analysis. Finally, facial expressions are learned from these trajectories using a variant of SVM, called ppfSVM, which allows to deal with the nonlinearity of the space of representations. Our experiments on four publicly available datasets showed that the proposed approach gives competitive or better than state-of-art results.  In the future, we will extend this approach to handle with smaller variations of facial expressions. Another direction could be adapting our approach for other applications that involve landmark sequences analysis such as action recognition. 

\section{Acknowledgements}
This work has been partially supported by PIA (ANR-11-EQPX-0023), European Founds for the Regional Development (FEDER-Presage 41779).


\end{document}